\documentclass{jlcl}

\usepackage{lineno}
\newcommand{\authornames}{Jonas Doumen\footnote{Faculty of Arts, KU Leuven, Leuven, Belgium \\Itec, imec research group at KU Leuven, Kortrijk, Belgium}, Veronica Juliana Schmalz$^1$, Katrien Beuls\footnote{Faculté d'informatique, Université de Namur, Namur, Belgium} and Paul Van Eecke$^{1,}$\footnote{Artificial Intelligence Laboratory, Vrije Universiteit Brussel, Brussels, Belgium}}

\newcommand{\authorlastnames}{Doumen, Schmalz, Beuls and Van Eecke}

\newcommand{\articletitle}{The Computational Learning of Construction Grammars:\\State of the Art and Prospective Roadmap}

\newcommand{\shortarticletitle}{Computational Models of Construction Grammar Learning}

\usepackage[utf8]{inputenc}
\usepackage[LFE,LAE,T1]{fontenc}
\usepackage{longtable, booktabs}
\usepackage{adjustbox}
\usepackage{graphicx}
\usepackage{rotating}
\usepackage{amsfonts}
\usepackage{amssymb}
\usepackage{mathtools}
\usepackage{textcomp}
\usepackage{graphicx}
\usepackage{longtable} % for 'longtable' environment
\usepackage{pdflscape} % for 'landscape' environment
\usepackage{enumitem}
\usepackage{hyperref}
\usepackage{parskip}
\usepackage{textcomp}

\newcolumntype{P}[2]{%
  >{\begin{turn}{#1}\begin{minipage}{#2}\small\raggedright\hspace{0pt}}l%
  <{\end{minipage}\end{turn}}%
}

\usepackage{framed}

\usepackage[]{natbib}

\begin{document}

\noindent
\begin{framed}
\noindent
\textcolor{blue}{\textit{Peer-reviewed author's draft of a journal article accepted for publication in Constructions and Frames (Vol. 17).}}
\end{framed}

\setcounter{page}{1}
\thispagestyle{plain}

\authordata

%%%%%%%%%%TODO%%%%%%%%%%%
% TEXT
% - Discussion & Conclusion

% TABLE
% - add groundedness as category
% - is everything from the text in the table? Bannard,
% - is everything that's in the table in the text?
% - is the table accurate?
% - update the table headers and values to reflect the ones introduced in the criteria

\section*{Abstract}

This paper documents and reviews the state of the art concerning computational models of construction grammar learning. It brings together prior work on the computational learning of form-meaning pairings, which has so far been studied in several distinct areas of research. The goal of this paper is threefold. First of all, it aims to synthesise the variety of methodologies that have been proposed to date and the results that have been obtained. Second,  it aims to identify those parts of the challenge that have been successfully tackled and reveal those that require further research. Finally, it aims to provide a roadmap which can help to boost and streamline future research efforts on the computational learning of large-scale, usage-based construction grammars.

\section{Learning Computational Construction Grammars}

The aim of this paper is to survey prior work on the computational learning of construction grammars, to identify gaps in the state of the art and to propose a perspective on the future of the field. The computational learning of construction grammars has traditionally been studied independently in different fields of research, in particular linguistics, cognitive science, computer science and artificial intelligence. As a consequence, research on this topic has been fragmented and interaction between the researchers involved has been scarce, as witnessed by a lack of cross-referencing. In order to address this issue, this paper brings together the variety of methodologies that have been proposed in the literature and synthesises the results that have been achieved to date. Specifically, we define 14 criteria in light of which we review 31 models of construction grammar learning. We then identify important gaps in the state of the art and propose a roadmap that can help streamline future research efforts and investments. Our hope is to inspire a new generation of construction grammarians to boost progress on scaling usage-based constructionist approaches to language.

On a high level, computational models for learning construction grammars are motivated by three main reasons. A first motivation is theoretical in nature and concerns the computational operationalisation and validation of usage-based theories of language acquisition. As usage-based theories of language argue that languages are learnt through their use in communication, computational models of language acquisition can contribute crucial evidence in favour or against the scalability of specific theoretical arguments \citep[see e.g.][]{chang2008constructing}. Second, the use of computational models brings important methodological advantages, as it enforces the use of precise and testable operational definitions, allows for the detection of theory-internal inconsistencies, and facilitates a fine-grained comparison of different theories. Such comparisons can play an important role in identifying inter-theoretical knowledge gaps and divergences \citep{bender2008grammar, mueller2015coregram}. Finally, the machine learning of construction grammars is important from an application and valorisation perspective, as it facilitates the use of construction grammar insights and analyses in broad-domain language technology applications \citep{vantrijp2022fcg}. Examples of such applications include visual question answering \citep[see e.g.][]{nevens2022language, verheyen2023neuro}, the frame-semantic analysis of discourse \citep[see e.g.][]{willaert2020building,beuls2021computational} and the construction-based analysis of corpora \citep[see e.g.][]{ehai2023sembrowse}. Apart from direct applications, computational models of language acquisition are also relevant to the broader fields of computational linguistics and artificial intelligence as they provide a method for intelligent agents to learn to communicate through protocols that exhibit the robustness, flexibility and adaptivity of human languages \citep{vaneecke2018generalisation, beuls2023fluid}.

The remainder of this paper is structured as follows. Section \ref{Methodology} presents the criteria that were used to include prior research efforts into this literature review, as well as the criteria in light of which these efforts are described in Section \ref{Prior Work}. Section \ref{Discussion} synthesises the state of the art, identifies challenges and opportunities, and proposes a roadmap for future research. Section \ref{Conclusion} concludes the paper.

\section{Methodology}
\label{Methodology}

\subsection{Inclusion Criteria}
\label{Inclusion Criteria}

The scope of this literature review concerns, to the best of our knowledge,  all published prior research on the computational learning of construction grammars, as defined through the following inclusion criteria: 

\begin{enumerate}
\item \textbf{Is the model concerned with learning form-meaning mappings?} This criterion concerns the nature of the linguistic knowledge that is learnt. Concretely, we only include models that concern the learning of form-meaning mappings. The selection is not influenced by the theoretical framework that is adopted nor by whether the authors situate their work within the field of construction grammar or not. 

\item \textbf{Is the model computationally operationalised?} The second criterion concerns the operationality of the learnt models. We only include models that have been implemented computationally.

\item \textbf{Can the model learn more than lexical mappings?} The third criterion concerns the nature of the constructions that are learnt. We only include models that acquire structures that transcend the level of individual words. For an overview of models that deal with vocabulary learning only, we refer the interested reader to \citet{krenn2020models}. 
\end{enumerate}

As research within this scope has been carried in a variety of fields of research, theoretical frameworks and scientific traditions, it was not feasible to design keyword searches or queries in scientific databases that would yield all relevant models. We came to the conclusion that any attempt to systematise the search process in such a manner left out important contributions that satisfy the inclusion criteria. Instead, we qualitatively scrutinised the literature to the best of our abilities and could include 31 models that satisfy the inclusion criteria.

\subsection{Discussion Criteria}
\label{Classification Criteria}

Prior work on learning construction grammars stems from different fields of research and therefore adopts a wide variety of methods, terminologies and experimental designs.
In order to streamline the discussion and facilitate a meaningful comparison, we introduce 14 discussion criteria that will guide the review of the included models.

\begin{enumerate}
\item \textbf{Learning task.} Which learning task does the model address? Which problem is the model designed to solve? Which evaluation criteria are used?
 
\item \textbf{Dataset.} To which datasets has the model been applied?

\item \textbf{Input.} What is the nature of the input to the model? 

\item \textbf{Form complexity.} What is the morpho-syntactic complexity of the linguistic input?

\item \textbf{Meaning complexity.} What is the semantic complexity of the linguistic input?

\item \textbf{Grounding.} Is the meaning representation grounded in a situation model?

\item \textbf{Segmentation level.} What level of segmentation of the input is provided to the model (phonemes, graphemes, words or utterances)?

\item \textbf{Lexicon.} Is a predefined lexicon provided?

\item \textbf{Grammatical categories.} Is a predefined set of grammatical categories provided?

\item \textbf{Incremental learning.} Does the model learn in an incremental fashion, i.e. dynamically extending its knowledge after each exemplar?

\item \textbf{Bi-directional grammar.} Can the learnt grammar be used for both language comprehension and production?

\item \textbf{Abstraction level.} What is the level of abstraction of the learnt constructions? Is the grammar limited to item-based constructions and slot-filling constructions or does it include hierarchical and recursive constructions?

\item \textbf{Non-compositionality.} Is the learning algorithm able to capture non-compositional aspects of language use? Is all meaning lexicalised or can grammatical constructions also contribute semantic information?

\item \textbf{Benchmark.} Is the data that was used during the learning and evaluation process precisely described and available to the community?

\end{enumerate}

Apart from guiding the literature review, the discussion criteria are also used to organise the schematic synthesis of the literature presented in Table \ref{tab:overview}. Here,  qualitative descriptions are provided for \textit{learning task}, \textit{dataset}, \textit{input}, \textit{form complexity} and \textit{meaning complexity. } Boolean values are used to assess the criteria of  \textit{grounding}, \textit{lexicon}, \textit{grammatical categories}, \textit{incremental learning}, \textit{bi-directional grammar}, \textit{non-compositionality} and \textit{benchmark}. \textit{Segmentation level }can either be `phonemes', `graphemes', or `words', and \textit{abstraction level} can either be `item-based',  or `hierarchical', with only the highest level of abstraction being listed.  The information contained in the table reflects the properties of the models as presented in the papers that introduce them, without implying that any limitations are inherent to the methodologies.

\section{Review of Prior Literature} 
\label{Prior Work}

This section provides a detailed review of all included models. A comparative analysis of all models in terms of the discussion criteria introduced in Section \ref{Classification Criteria} is also provided in Table \ref{tab:overview}. Readers who are primarily interested in the higher-level picture can safely skip ahead to Section \ref{Discussion}, where we present a synthesis of the details covered in this section. 

On the highest level, the models can be organised according to the overall learning task that they are designed to tackle:

\begin{enumerate}

\item \textbf{Learning a maximally concise grammar.} The task is to find a minimal set of constructions that optimally covers a corpus of language use.

\item \textbf{Learning a grammar from utterance-meaning pairs.} The task is to learn a grammar that maps between utterances and their meaning representation, whereby a gold semantic annotation is provided for each utterance.

\item \textbf{Learning a grammar under referential uncertainty.} The task is to learn a grammar that maps between utterances and their meaning representation, whereby a superset of the gold semantic annotation is provided. The referential uncertainty stems from the fact that the exact meaning representation is not provided.
 
\item \textbf{Learning a grammar from a situation model.} The task is to learn a grammar that maps between utterances and their meaning representation, whereby no gold semantic annotation is provided. The meaning has to be abductively derived from a situation model.
\end{enumerate}

The following sections reflect the organisation of the models in terms of these four learning tasks.

%%%%SUMMARY TABLE 

\begin{landscape}

\begin{tiny}
\begin{longtable}{p{1.3cm}|p{1.3cm}|p{1.3cm}|p{1.3cm}|p{.9cm}|p{.9cm}|c|c|c|c |c |c |p{1cm} |c |c }

\caption{Comparative analysis of the literature in terms of the discussion criteria proposed in Section \ref{Classification Criteria}. } \label{tab:overview} \\   \toprule
  & 
  \multicolumn{1}{P{90}{1.8cm}|}{\textbf{\tiny Learning Task}} &
  \multicolumn{1}{P{90}{1.8cm}|}{\textbf{\tiny Dataset}} &
  \multicolumn{1}{P{90}{1.8cm}|}{\textbf{\tiny Input}} &
  \multicolumn{1}{P{90}{1.8cm}@{}|}{\textbf{\tiny Form complexity}} &
  \multicolumn{1}{P{90}{1.8cm}@{}|}{\textbf{\tiny Semantic complexity}} &
    \multicolumn{1}{P{90}{1.8cm}@{}|}{\textbf{\tiny Grounded}} &
  \multicolumn{1}{P{90}{1.8cm}@{}|}{\textbf{\tiny Lexicon given}} & 
  \multicolumn{1}{P{90}{2cm}@{}|}{\textbf{\tiny Categories given}} & 
  \multicolumn{1}{P{90}{1.8cm}@{}|}{\textbf{\tiny Segmentation level}} & 
  \multicolumn{1}{P{90}{1.8cm}@{}|}{\textbf{\tiny Incremental}} &
  \multicolumn{1}{P{90}{1.8cm}@{}|}{\textbf{\tiny Bi-directional}} &   
  \multicolumn{1}{P{90}{1.8cm}@{}|}{\textbf{\tiny Abstraction level}} &
  \multicolumn{1}{P{90}{1.8cm}@{}|}{\textbf{\tiny Non-compositionality}}&
  \multicolumn{1}{P{90}{1.8cm}@{}}{\textbf{\tiny Benchmark}} \\
 \midrule
 \endhead
  \cite{abend2017bootstrapping} & learning a grammar under referential uncertainty & CHILDES Eve section, with automatically generated meaning representations & utterances with candidate meaning representations & simple utterances (intransitives and transitives)& one predicate per utterance & N & N & Y & words & Y & N & hierarchy & N & N \\
 \midrule
  \cite{alishahi2008computational} &  learning a grammar from utterance-meaning pairs & English synthetic data & utterance-meaning pairs & simple utterances (intransitives,  transitives,  ditransitives) & one predicate per utterance & N & Y & Y & words & Y & Y & item-based & N & N \\
   \midrule
\cite{artzi2013weakly} & learning a grammar from a situation model & Navi corpus  & utterance, start state and validation function & navigation instructions with spatial relations  & lambda calculus with Neo-Davidsonian event semantics & Y & Y  & Y & words & N & N & hierarchy & N & Y \\
 \midrule
  \cite{beekhuizen2014automating} & learning a grammar under referential uncertainty & English synthetic data & utterances with candidate meaning representations  & simple utterances (intransitives and transitives) & one predicate per utterance  & N & N  & Y & words & Y & N & hierarchy & Y & N \\
 \midrule 
  \cite{beekhuizen2015constructions} & learning a grammar under referential uncertainty &  English synthetic data & utterances with candidate meaning representations & simple utterances (intransitives and transitives) & one predicate per utterance  & N & N & Y & words & Y & Y & hierarchy & Y & N \\
 \midrule 
  \cite{beuls2010situated} &  learning a grammar from utterance-meaning pairs & Hungarian synthetic data & utterance-meaning pairs & simple utterances with poly-personal verbal agreement & one predicate per utterance  & N & Y & Y & words &  Y &  Y & item-based & Y & N \\
 \midrule 
  \cite{chang2008constructing} & learning a grammar from utterance-meaning pairs & CHILDES Naomi section with manually annotated meaning representations & utterance-meaning pairs & simple utterances (intransitives and transitives) & one predicate per utterance  & N & Y & Y & words & Y & N & item-based & Y & N \\
 \midrule 
  \cite{chen2008learning} & learning a grammar under referential uncertainty & RoboCup & utterances with candidate meaning representations & simple utterances (intransitives and transitives) & one predicate per utterance  & Y & N & N & words & N & Y & unk. & N & Y\\
 \midrule
  \cite{dominey2005emergence} & learning a grammar from utterance-meaning pairs & English synthetic data & utterance-meaning pairs (derived from object and event detection) & simple utterances (intransitives,  transitives,  ditransitives) & max. two predicates per utterance & Y & N & Y & words & Y & N & item-based & N & N \\
 \midrule
 \cite{dominey2005learning} &  learning a grammar from utterance-meaning pairs & English synthetic data  & utterance-meaning pairs (derived from object and event detection) & simple utterances (intransitives,  transitives,  ditransitives) & one predicate per utterance & Y & N & Y & words & N & Y &item-based  & N & N\\
 \midrule 
 \cite{doumen2023modelling} & learning a grammar from utterance-meaning pairs & CLEVR & utterance-meaning pairs & simple queries about properties of objects & procedural semantic networks & N & N & N & words & Y & Y &item-based  & Y & Y\\
 \midrule 
 \cite{dunn2017computational} & learning a maximally concise grammar & English web data from ukWac & POS and semantic tags & full (broad coverage) & semantic tags & N & Y & Y & words & N & N & hierarchy & Y & Y \\
\midrule
 \cite{dunn2018modeling} & learning a maximally concise grammar & multilingual web data from WAC and Aranea& POS and semantic tags & full (broad coverage) & semantic clusters & N & N & Y & words & N & N & hierarchy & Y & Y \\
  \midrule 
   \cite{garcia2015usage} &  learning a grammar from utterance-meaning pairs & English synthetic data &  utterance-meaning pairs & simple utterances with recursion & semantic networks in second-order logic & Y & Y & N & words &  Y &  Y  & hierarchy & Y & N \\
\midrule
\cite{garcia2016insight} &  learning a grammar from utterance-meaning pairs & English synthetic data & utterance-meaning pairs & simple utterances with recursion & semantic networks in second-order logic & Y & Y & N & words &  Y &  Y  & hierarchy & Y &N  \\
  \midrule
  \cite{garcia2018origins} &  learning a grammar from utterance-meaning pairs  & English synthetic data & utterance-meaning pairs & simple utterances with recursion & semantic networks in second-order logic & Y &Y & N & words &  Y &  Y  & hierarchy & Y &N  \\
  \midrule
 \cite{gaspers2011unsupervised} & learning a grammar under referential uncertainty & RoboCup & utterances with candidate meaning representations & simple utterances (intransitives and transitives) & one predicate per utterance & N & N & N & words & N & N & item-based & unk. & Y\\
 \midrule
 \cite{gaspers2012usage} & learning a grammar under referential uncertainty & RoboCup & utterances with candidate meaning representations  & simple utterances (intransitives and transitives) & one predicate per utterance & N  & N & N & phon.  & Y & N & item-based & unk. & Y \\
 \midrule 
  \cite{gaspers2014computational} & learning a grammar under referential uncertainty & RoboCup & utterances with candidate meaning representations  &  simple utterances (intransitives and transitives) & one predicate per utterance & N  & N & N & phon.  & Y & Y & item-based  & unk. & Y\\
 \midrule 
 \cite{gaspers2016constructing} & learning a grammar under referential uncertainty & RoboCup & utterances with candidate meaning representations & simple utterances (intransitives and transitives) & one predicate per utterance & N  & N & N & phon.  &  Y & N & item-based & unk. & Y \\
 \midrule
  \cite{gerasymova2010acquisition} &  learning a grammar from utterance-meaning pairs & Russian synthetic data & utterance-meaning pairs & simple utterances with aspectual distinctions & procedural semantic networks & N &  Y & Y & words &  Y &  Y & item-based  & Y & N\\
 \midrule 
 \cite{gerasymova2012experiment} & learning a grammar from utterance-meaning pairs & Russian synthetic data & utterance-meaning pairs & simple utterances with aspectual distinctions & procedural semantic networks & N & Y & Y & words &  Y &  Y  & item-based  & Y &  N\\
 \midrule 
 \cite{kwiatkowski2010inducing} & learning a grammar from utterance-meaning pairs & multilingual GeoQuery dataset  & utterance-meaning pairs & complex queries with recursion, disjunction and conjunction & typed lambda calculus & N & N & Y & words & N & N & hierarchy &  N & Y \\
 \midrule
  \cite{kwiatkowski2011lexical} & learning a grammar from utterance-meaning pairs & multilingual GeoQuery dataset and ATIS  & utterance-meaning pairs & complex queries with recursion, disjunction and conjunction & typed lambda calculus & N & N & Y & words & N & N & hierarchy &  N & Y \\
 \midrule
  \cite{kwiatkowski2012probabilistic} & learning a grammar under referential uncertainty & CHILDES Eve section, with automatically generated meaning representations & utterances with candidate meaning representations & simple utterances (intransitives and transitives) & one predicate per utterance & N & N & Y & words & Y & N & hierarchy &  N & N \\
 \midrule
  \cite{marti2021discover} & learning a maximally concise grammar & Spanish web data & POS and dependency labels & full (broad coverage) & semantic tags & N & Y & Y & words & N & N & hierarchy & Y & Y \\
\midrule
  \cite{nevens2022language} & learning a grammar from a situation model & CLEVR & questions and feedback & simple queries about properties of objects & procedural semantic networks & Y & N & N & words & Y & Y &item-based  & Y & Y\\
 \midrule 
 \cite{spranger2015acquisition} & learning a grammar from a situation model & English synthetic data & questions and feedback & simple utterances expressing spatial relations & procedural semantic networks & Y & Y & Y & words &  Y &  Y  & item-based  & Y & N \\
 \midrule 
 \cite{spranger2015incremental} & learning a grammar from a situation model  &  English synthetic data & questions and feedback & simple utterances expressing spatial relations & procedural semantic networks & Y &  Y & Y & words &  Y &  Y  & item-based  & Y & N\\
 \midrule 
 \cite{spranger2017usage} & learning a grammar from a situation model & English synthetic data &  questions and feedback & simple utterances expressing spatial relations & procedural semantic networks & Y & Y & Y & words &  Y &  Y  & item-based & Y & N \\
  \midrule 
 \cite{steels2004constructivist} &  learning a grammar from utterance-meaning pairs & English synthetic data & utterance-meaning pairs (derived from object and event detection) & simple utterances & predicate calculus & Y &  N & N & words &  Y &  Y & hierarchy & Y & N\\
\bottomrule

\end{longtable}
\end{tiny}

\end{landscape}

\subsection{Learning a maximally concise grammar}
\label{concise grammar}

The first category of models addresses the task of finding a minimal set of constructions that optimally covers a corpus of language use. \citet{dunn2017computational} introduces a method to induce schematic patterns from large amounts of web-crawled corpus data. These patterns take the form of a sequence of slots that can be filled by word forms, morpho-syntactic categories or semantic categories. These categories are provided in the form of annotation layers. The resulting grammars are evaluated against a held-out test set in terms of various measures, including minimal description length and $\Delta P$ entrenchment \citep{ellis2006language}. Variations on the method are described by \citet{dunn2018modeling, dunn2019frequency,dunn2022exposure, dunn2023exploring} and \citet{dunn2021learned}. \citet{marti2021discover} present DISCO, a methodology for discovering constructional candidates in large web-crawled corpora of Spanish texts. Similar to \citet{dunn2017computational}, the semantic categories integrated into the `lexico-syntactic' patterns are modelled through as clusters over the distributional semantic representations of lemmata. The retrieved patterns are evaluated using statistical association measures as well as through manual evaluation by expert linguists. %\citet{barteld2020construction} explore a similar method for the identification of construction candidates in German.

This line of work models the induction of partially abstract patterns that combine form-related and meaning-related features. While relevant from a construction grammar perspective, the resulting patterns do not correspond to constructions that actually constitute mappings between aspects of form and meaning. These models thereby do not support mapping between utterances and their meaning representations.

\subsection{Learning a grammar from utterance-meaning pairs}
\label{predicate-arguments}

The second category of models addresses the task of learning a construction grammar from utterance-meaning pairs.

\citet{dominey2005emergence,dominey2005sensorimotor,dominey2006holophrases} and \citet{dominey2005learning} present a neural model for the acquisition of holophrase constructions, item-based constructions and abstract constructions that capture argument structure relations (e.g.  transitives and ditransitives).  Learners start with the capability to distinguish between closed-class and open-class words and learn to map between slots in the argument structure constructions and the semantic roles they take. Learning the meaning of open-class words is tackled as an initial cross-situational learning step. Item-based constructions are then learned from the remaining closed-class words (e.g. \textit{X was Y to Z by A}) by storing the mapping of the order of  thematic roles (e.g. object, action, recipient, agent) and these closed-class elements as constructions in the construction inventory. 

\citet{alishahi2008computational} present a computational model that mimics the acquisition of verb argument structure. %learning here
The input utterance-meaning pairs are generated based on the 20 most frequent verbs of a subsection of the CHILDES corpus \citep{macwhinney2000childes}. An example would be the utterance ``\textit{Mom put toys in boxes}'' paired to the meaning representation \textsc{put\textsubscript{[cause,move]}(mom\textsubscript{ \textlangle agent\textrangle}, toys\textsubscript{ \textlangle theme\textrangle}, in\textsubscript{[]}(boxes\textsubscript{\textlangle destination\textrangle})\textsubscript{\textlangle destination\textrangle})}. The lexicon and semantic roles are given to the model a priori. A construction is defined as a probabilistic association between syntactic and semantic features of a verb and its arguments, which emerged over a number of observations. Language processing is tackled as a Bayesian prediction problem. In production, the model predicts the syntax from a given semantic structure, and in comprehension, it predicts a (partial) semantic structure for a given utterance. The authors present experiments for both comprehension and production. The experiments show that the model exhibits effects of syntactic and semantic bootstrapping, that it can recover from over-generalisation, and that it is robust to noise, thereby mimicking the stages of child language acquisition.

\citet{chang2008constructing} presents a set of learning operators that operationalise Bayesian model merging in the framework of Embodied Construction Grammar (ECG). The task is to learn a grammar that can correctly comprehend utterances from a held-out test set at certain intervals during training. The dataset that is used consists in a schema-annotated subset of the CHILDES corpus, containing parent-child interactions from a child from 15 until 24 months of age. A lexicon and ontology are provided, corresponding to the linguistic knowledge of a child at the two-word stage. Two classes of learning operators are introduced: mapping operators and reorganisation operators. In the first class of operators, the simple mapping operator maps either all, or a part of the uncovered form to its co-occurring context information. Relational mapping creates a new construction that encodes the syntactic and semantic relations between existing constructions that could apply. The applied constructions become constituents in the newly formed item-based construction. For the input utterance ``\textit{throw ball}'' for example, the lexical constructions for  \textsc{throw} and \textsc{ball} may already be known. A new construction is then created that encodes that \textsc{throw} precedes \textsc{ball}, and that the ball is the throwee in the throwing event. In the class of reorganisation operators, a structural alignment step pre-computes compatible alignments for further reorganisation. Next, three operators can apply: merging, joining and splitting. The merge operator learns a generalised construction for two constructions that contain shared structure. For example, \textsc{throw-ball} and \textsc{throw-block} can be generalised to \textsc{throw-toy}. A second reorganisation operator joins existing constructions that have a certain overlap, for example \textsc{human-throw} and \textsc{throw-bottle} can be joined into \textsc{human-throw-bottle}. Inversely, the splitting operator creates new constructions by taking the set difference between existing constructions, for example \textsc{throw-frisbee} and \textsc{throw} can be split into \textsc{frisbee}, and a new item-based construction, \textsc{throw-X}, which takes the newly created \textsc{frisbee}-construction as a constituent. After the application of each applicable learning operator for a given input, the model uses minimum description length to select the shortest grammar that covers the data. The model is evaluated in relation to a hand-crafted gold standard grammar in terms of form and meaning coverage, and description length. The presented experiments operate at a rather small scale: first on 40 tokens of the verbs fall and throw, then on 200 tokens of caused-motion and self-motion predicates. The author reports that, in qualitative evaluation, both experiments show a resemblance to the stages described in the literature of usage-based language acquisition, i.e. going from concrete to increasingly abstract constructions.

\citet{kwiatkowski2010inducing} introduce a unification-based learning algorithm, which is designed to induce semantic parsers from corpora consisting of sentence and logical-form pairs. The model learns a lexicon and parameters in Combinatory Categorial Grammar (CCG). In CCG, a fixed set of combinatory rules is given, which govern how both categories and arguments of logical predicates can be combined. A lexical entry consists of a word form, its syntactic category and a logical predicate, for example: \textit{what \(\vdash\) S / NP : \(\lambda\)x.answer(x)}. The model initially creates a holistic construction by mapping an entire sentence to its logical representation. The main generalisation operator splits these overly specific entries into smaller, more widely applicable lexical entries. Higher-order unification is used to restrict the number of hypotheses, while safeguarding the integrity and combinatory properties of the logical forms. As there are many possible analyses for a given sentence, a log-linear model is used to select the most likely analysis. Before training, the model is initialised with a list of NPs that are present in the data (e.g. \textit{Texas \(\vdash\) NP : tex}). The model is evaluated for four languages and two semantic representation formats on the GeoQuery dataset \citep{zelle1996learning}. \citet{kwiatkowski2011lexical} apply a further iteration of the same methodology to the more natural Air Travel Information System pilot corpus \citep[ATIS --][]{hemphill1990atis}.
In contrast to the other work presented in this section, CCG assumes that meaning is fully lexicalised, and thereby that language is fully compositional.

\citet{gerasymova2010acquisition,gerasymova2012experiment} investigate the acquisition of holophrase constructions, item-based constructions and abstract constructions, represented and processed using the Fluid Construction Grammar (FCG) framework \citep{steels2006unify, vantrijp2022fcg, beuls2023fluid}, for Russian aspectual marking in a tutor-learner language game setting \citep{steels1998origins, steels2001language}.  Holophrase constructions are learnt by a straightforward mapping operation between an observed form and its meaning.  Item-based constructions and abstract constructions are learnt as generalisations over pre-categorised lexical items.  \citet{beuls2010situated} apply the same methodology to the conjugation of verbs in Hungarian,  with a special focus on its intricate agreement marking system. \citet{garcia2015usage, garcia2016insight} and \citet{garcia2018origins} apply the same paradigm to the acquisition of abstract, hierarchical and recursive constructions. 

%\citet{artzi2015broad} introduces a broad-coverage CCG-driven semantic parser for the AMR Bank \cite{banarescu2013abstract}. The authors introduce globally scoped Skolem terms and a factor graph methodology to handle non-compositional reasoning, such as anaphoric dependencies. A seed lexicon, NER and POS tags, and CCGBank categories are provided a priori. The presented \textsc{Smatch} test results set a new state of the art for the AMR Bank, while  outperforming earlier dependency-tree-based approaches.

\citet{doumen2023modelling} present a model of the co-acquisition of grammatical categories and form-meaning mappings, ranging from lexical to item-based constructions. The model operationalises \citet{tomasello2003constructing}'s pattern finding mechanisms as a set of meta-layer learning operators in Fluid Construction Grammar. The base operator stores an entire utterance with its meaning representation as a holophrase construction. A second class of operators generalises over holophrase constructions and new observations. When a difference in both form and meaning is found, these operators learn the syntactic and semantic composition of the observation by means of substitution, addition and deletion operations, resulting in item-based constructions, lexical constructions and a network of emergent grammatical categories that governs how constructions can combine. A third class of operators starts from partial analyses. When an observation is partially covered by one or multiple constructions in the agent's inventory, the model applies these to the utterance, and a new item-based or lexical construction  is learned, along with corresponding categorial links. A final learning operator adds new categorial links if all constructions that are necessary to cover an utterance are known, but no links in the categorial network are found. The model is evaluated on a subset of the CLEVR dataset \citep{johnson2017clevr}.

\subsection{Learning a grammar under referential uncertainty}
\label{uncertainty}

The third category of models addresses the task of learning a construction grammar under referential uncertainty. While the previous category of models took as input utterances paired with their semantic representation, this category of models takes as input utterances paired with a superset of their semantic representation. The fact that the meaning representation is noisy thereby poses an additional challenge to the learning task.

A first line of work in this category focuses on aligning commentaries with observed actions in Robocup football games, a task introduced by \citet{chen2008learning}. The input data concerns a list of natural language utterances of limited morpho-syntactic complexity, each paired with a number of candidate meaning representations. The correct meaning representation for an utterance always consists in a single predicate. The task is then to learn a productive grammar that can map between utterances and their meaning representation. Along with the task and corpus, \citet{chen2008learning} also present a method to induce \textit{probabilistic synchronous context-free grammars} that can effectively be used for semantic parsing and meaning-driven language production. \citet{gaspers2011unsupervised} make use of the same task and corpus to develop and evaluate a method to learn construction grammars without strict supervision. In a first phase, a probabilistic lexicon is learned using a cross-situational learning algorithm \citep{fazly2010probabilistic}. The lexicon is restricted to sequences of one or two tokens, paired with one predicate or argument that occurs in the corpus. In a second phase, schemata are computed as generalisations over co-occurring lexical items and predicates or arguments. In particular, lexical items appearing in the same syntactic environments are grouped into \textit{semantic equivalence classes}. Then, schemata are constructed by substituting the lexical items with abstract labels associated to their respective semantic equivalence class. \citet{gaspers2012usage, gaspers2014computational} and \citet{gaspers2016constructing} present variations on this model, where the level of segmentation of the input is reduced from tokens through graphemes to phonemes.

Introducing the framework of Meaningful Unsupervised Data Oriented Parsing ($\mu$-DOP), \citet{beekhuizen2014automating} present a model that incrementally learns mappings between nodes in binary trees and meaning representations expressed in predicate logic. The task that the model aims to solve is to find the correct predicate-argument structure for a given utterance in referentially ambiguous scenes. They apply the model to artificial data that is not fully compositional on the meaning side, including for example idioms. At the start of an experiment, the construction inventory is empty. When presented with a new utterance, the model first attempts to apply any existing constructions. In order to induce syntactic structures, the model first generates all possible binary trees for a given utterance. Similarly, on the meaning side, the model decomposes the meaning predicates into all possible decompositions. The arguments are replaced by slot indices. For example \textsc{watch(E1, E2)} can be further decomposed into \textsc{P(E1, E2)} and \textsc{P : watch}. 
 All unanalysed nodes in the binary trees are then paired with all possible decompositions of parts of the semantic representation as candidate form-meaning pairs. These are then combined with the possible binary tree analyses to form a hypothesis space of possible analyses. As a last step, the model proceeds to extract all subtrees, and compares them to previously observed subtrees. If a subtree is found in multiple utterances' parse trees, then its prior probability of being a valid constituent increases. Non-compositional parts are retained by only allowing further decomposition at nodes where a meaning representation is present. In both parsing and production, the derivation with the highest joint prior probability is selected, which in practice means that hypotheses with a smaller amount of subtrees are preferred. As a further evolution of this work, \citet{beekhuizen2014usage} and \citet{beekhuizen2015constructions} introduce the Syntagmatic-Paradigmatic Learner (SPL), which addresses a number of desiderata inspired by usage-based theories of language acquisition. As such, the generation of possible analyses, called derivations, is driven by a set of general structure-building operations, implementing parts-to-whole learning strategies. These operations allow for the concatenation of derivations, slot filling, syntactic bootstrapping, and ignoring words. The model is now applied to artificial data that is generated based on empirical research on noise, uncertainty, and situational continuation in child-directed communicative interactions.

\citet{kwiatkowski2012probabilistic} and \citet{abend2017bootstrapping} introduce a probabilistic learning algorithm for combinatory categorial grammars (CCG), designed to model the human language acquisition process. This approach differs from earlier work by \citet{kwiatkowski2010inducing}, presented in Section \ref{predicate-arguments}, in a few key aspects. An evident difference is that this model learns from utterance-situation pairs under propositional uncertainty, rather than having a single gold standard semantic representation for each utterance. A second difference is that the model is able to learn the lexicon, and the mapping between the lexicon and predefined categories from scratch, rather than relying on a list of noun phrases. An important assumption in this line of work is that a language-independent set of combinatory rules is provided to the model. The grammar is evaluated both quantitatively and qualitatively using the Eve corpus \citep{brown1973first}, which was enriched with deterministically mapped logical forms based on dependency tree annotations.

\subsection{Learning a grammar from a situation model}
The final category of models addresses the task of learning a construction grammar from utterances observed during communicative interactions. As such, the meaning representation of the utterances is not provided, but needs to be abductively inferred from the situation in which the interaction takes place. %the emergence of complex semantic structures

\citet{steels2004constructivist} presents an initial experiment in which a population of artificial agents bootstrap argument structure constructions and grammatical categories from visually grounded meaning representations. The paper presents a description game, in which two agents from the population observe a scene that needs to be successfully described by one agent to the other. The scene is transcribed in terms of first order logic predicates. During conceptualisation, the speaker agent decides on the `event profile' that they want to express \citep[see e.g.][]{croft1998event}, i.c. deciding which roles have to be expressed linguistically. Whether speaking or listening, an agent first makes use of its own construction grammar to process the conceptualised meaning representation or observed utterance. When an agent fails to formulate an utterance that expresses the conceptualised meaning representation or fails to comprehend an observed utterance in terms of the current scene, a game fails and a learning event takes place. When a game succeeds, the score of an agent's applied constructions is increased, while the score of competing constructions is decreased. Upon failure, the scores of the constructions that were used are decreased. Overall, the presented methodology shows how hierarchical semantic and syntactic categories can be learnt, and how the emergence of syntax aids to resolve ambiguity. \cite{vantrijp2008analogy,vantrijp2016evolution} presents an extensive suite of follow-up experiments and shows how abstract semantic roles can emerge and evolve in populations of autonomous agents through multi-level selection strategies.

\citet{artzi2013weakly} present a model where a seed lexicon, a predefined set of combinatory rules and a situation model are provided. The seed lexicon and combinatory rules are used to generate hypotheses about the meaning underlying observed utterances, which can subsequently be validated against the situation model. The goal is to extend the seed lexicon with new items in order to solve a navigation task. 

\citet{spranger2015acquisition} and \citet{spranger2015incremental, spranger2017usage} present a model of the acquisition of spatial language in embodied artificial agents. In a shared environment and through a tutor-learner language game, two agents interact with 15 different objects in over 1000 different spatial scenes. The goal is to simultaneously acquire the semantic and syntactic aspects of spatial language. Each communicative interaction proceeds as follows. The tutor agent selects formulates an utterance that uniquely refers to an object in the situational context. The learner agent comprehends and interprets the utterance with respect to the scene and points to the object that results from the interpretation process. Then, feedback is provided by the tutor in the form of pointing. If the learner agent misinterpreted the observed utterance, it needs to make a hypothesis about the intended meaning of the utterance. The agent does this by composing a procedural semantic network based on a set of cognitive operations that it can perform. The observed form and hypothesised meaning can then be stored in the form of a holophrastic construction. Later, the agent can generalise over the constructions it knows and thereby construct more abstract constructions. The semantic classes of the slots in the more abstract constructions are predefined in an ontology that the learner can access.

\citet{nevens2022language} build further on the composition processes introduced by \citet{spranger2015acquisition} and the syntactico-semantic generalisation operators introduced by \citet{doumen2023modelling}. They tackle the visual question answering benchmark introduced by \citet{johnson2017clevr} and are able to bootstrap a construction grammar that maps between English questions and visually grounded queries without ever having observed the queries. Through lateral inhibition alignment dynamics, generally applicable constructions become gradually more entrenched, while suboptimal constructions, i.e. constructions that do not generalise to other scenes, for example due to suboptimal hypotheses about the intended meaning of an observed utterance, gradually disappear from the grammar.

\section{Discussion}
\label{Discussion}

The low-level review of the prior literature in the previous section reveals perhaps most clearly that existing models for computationally learning construction grammars are highly diverse in nature and therefore challenging to compare. Their diversity is situated on almost any level, from the task that is tackled, to the goals that are envisioned, the datasets that are used, the approaches that are taken and the methodologies that are applied. So, where are we at now and where should we be heading?

Answering these questions presupposes that a particular perspective is taken. After all, researchers in natural language processing, for example, have different immediate goals and concerns than researchers in cognitive science or language pedagogy. We will be addressing these questions from a constructionist perspective, with the general goal in mind of operationalising large-scale, usage-based construction grammar. So, what does it mean to operationalise usage-based construction grammar on a large scale? We define the term \textit{operational} as having a computational implementation that supports the processes of language comprehension (i.e. mapping from an utterance to a representation of its meaning) and production (i.e. mapping from a meaning representation to an utterance that expresses it). The term \textit{large-scale} can be interpreted as having a broad and domain-general coverage of the language. For the term \textit{usage-based}, we adhere to \citet{bybee2006usage}'s view that an individual's grammar is rooted in and shaped by the individual's history of communicative interactions. Finally, we interpret the term \textit{construction grammar} as adhering to the basic principles underlying constructionist approaches to language \citep{goldberg2003constructions}. These principles are summarised by \citet{vantrijp2022fcg} as follows: (i) all linguistic knowledge is captured in the form of constructions, i.e. form-meaning pairings, (ii) there is no strict distinction between words and grammar rules, so that non-compositional aspects of the language can elegantly be captured, (iii) constructions cut across the different levels of linguistic analysis, and (iv) construction grammars are learnt, dynamic systems. 

An important insight gained through the literature review is that prior models typically score well on one or two of these dimensions, while suffering from substantial shortcomings on the others. As such, the models that address the \textit{large-scale} dimension of the challenge, i.e. those by \citet{dunn2017computational,dunn2018modeling,dunn2019frequency,dunn2022exposure,dunn2023exploring}, \citet{dunn2021learned} and \citet{marti2021discover}, do not yield individual grammars that support language comprehension and production. The scale of other models is restricted in one of three ways. Some models cover specific linguistic phenomena only, such as basic argument structure \citep{steels2004constructivist,dominey2005emergence,dominey2005sensorimotor,
dominey2005learning,dominey2006holophrases,vantrijp2008analogy, vantrijp2016evolution, garcia2015usage,garcia2016insight,garcia2018origins}, aspectual marking \citep{gerasymova2010acquisition,gerasymova2012experiment}, agreement marking \citep{beuls2010situated} or spatial relations \citep{spranger2015acquisition,spranger2015incremental,spranger2017usage}. Other models are applied to narrow artificial datasets, such as Robocup \citep{chen2008learning,gaspers2011unsupervised,gaspers2012usage,gaspers2014computational,gaspers2016constructing}, GeoQuery \citep{kwiatkowski2010inducing,kwiatkowski2011lexical}, CLEVR \citep{nevens2022language,doumen2023modelling}, ATIS \citep{kwiatkowski2011lexical}, Navi \citep{artzi2013weakly}, or other datasets specifically generated for evaluating the model \citep{beekhuizen2014automating,beekhuizen2014usage,beekhuizen2015constructions}. A final set of models makes use of small-scale corpora of transcribed children's speech \citep{alishahi2008computational,chang2008constructing,kwiatkowski2012probabilistic,abend2017bootstrapping}. The large-scale operationalisation of construction grammar learning thereby remains very much an open challenge.

When it comes to the \textit{usage-based} nature of the models, we can define a broad spectrum of possible approaches. On the most usage-based side of the spectrum, we would expect to find models that learn individual construction grammars from empirically observed communicative interactions, without access to a segmentation of the utterances, their meaning representations, or any predefined categories or lexical items. No models that satisfy all of these criteria were found in the literature. In fact, all prior models learn from utterances that are segmented on some level (phonemes, graphemes or words). Existing models that learn individual grammars without any access to annotated meaning representations are always learnt based on artificial utterances \citep{nevens2022language}, sometimes in combination with a predefined lexicon and system of categories \citep{artzi2013weakly, spranger2015acquisition,spranger2015incremental,spranger2017usage}. Other models that do not have access to exact meaning representations do have access to a superset of the meaning representations. These models learn from artificial utterances \citep{beekhuizen2014automating,beekhuizen2014usage,beekhuizen2015constructions} or short natural utterances \citep{chen2008learning,gaspers2011unsupervised,gaspers2012usage,gaspers2014computational,gaspers2016constructing}, sometimes in combination with a set of predefined combinatory rules \citep{kwiatkowski2012probabilistic,abend2017bootstrapping}. Next, we find a class of models that learn from semantically annotated utterances. Some of these models learn construction grammars based on artificial data \citep{doumen2023modelling}, almost always augmented with with a pre-defined lexicon \citep{steels2004constructivist,vantrijp2008analogy, vantrijp2016evolution, garcia2015usage,garcia2016insight,garcia2018origins}, pre-categorised lexical items \citep{dominey2005emergence,dominey2005sensorimotor,dominey2005learning,dominey2006holophrases,beuls2010situated,gerasymova2010acquisition,
gerasymova2012experiment} or a predefined set of categories and combinatory rules \citep{kwiatkowski2010inducing}.  Other models in this class learn from natural data augmented with pre-categorised lexical items \citep{alishahi2008computational,chang2008constructing} or a predefined set of categories and combinatory rules \citep{kwiatkowski2011lexical}. A final class of models learns from natural data enhanced with morpho-syntactic and semantic annotation layers \citep{dunn2017computational,dunn2018modeling,dunn2019frequency,dunn2021learned,marti2021discover,dunn2022exposure,dunn2023exploring}. However, these models do not yield grammars that support language comprehension and production. 

The third dimension concerns the degree to which the models adhere to the basic tenets of \textit{construction grammar}. Fully constructionist approaches should support bi-directional language processing. They should capture all linguistic knowledge in the form of form-meaning pairings, allow for meaningful schemata above the word level, allow for combining information from different levels of linguistic analysis, and support incremental, individual and adaptive learning. On the most constructionist side of this spectrum, we find approaches that explicitly make use of computational construction grammar implementations to represent and process linguistic structures. Models adopting Fluid Construction Grammar \citep{steels2004constructivist,vantrijp2008analogy,beuls2010situated,gerasymova2010acquisition,
gerasymova2012experiment,spranger2015acquisition,spranger2015incremental,vantrijp2016evolution,spranger2017usage,garcia2015usage,
garcia2016insight,garcia2018origins,nevens2022language,doumen2023modelling} faithfully adhere to the basic tenets of construction grammar. Models adopting $\mu$-DOP \citep{beekhuizen2014automating,beekhuizen2014usage,beekhuizen2015constructions} capture form-meaning pairings beyond the level of individual words and support language comprehension and production, but are inherently word order-based. \citet{alishahi2008computational}, \citet{gaspers2011unsupervised,gaspers2016constructing} and \citet{gaspers2012usage,gaspers2014computational} introduce computational operationalisations of construction grammar that are specific for the purposes of the models they describe. In essence, these operationalisations adhere to the basic principles of construction grammar, but they have only been operationalised for basic word order-based argument structure patterns. Models adopting Embodied Construction Grammar \citep{chang2008constructing} are constructionist in nature but only support language processing in the comprehension direction. Finally, in the case of models based on Dynamic Construction Grammar \citep{dominey2005emergence,dominey2005sensorimotor,dominey2005learning,dominey2006holophrases}, it is difficult to tell inhowfar they capture form-meaning pairings of varying degrees of abstraction, in particular when it comes to modelling phenomena that do not depend on sequential word order. A second class of models makes use of generative grammar formalisms and thereby assumes that the meaning of linguistic expressions mirrors the compositional structure of their syntactic analysis. These models make use of Combinatory Categorial Grammar \citep{kwiatkowski2010inducing,kwiatkowski2011lexical,kwiatkowski2012probabilistic,artzi2013weakly,abend2017bootstrapping} or probabilistic context-free grammar \citep{chen2008learning} as the underlying linguistic framework. Finally, the models by \citet{dunn2017computational,dunn2018modeling,dunn2019frequency,dunn2022exposure,dunn2023exploring}, \citet{marti2021discover} and \citet{dunn2021learned} capture partially abstract word order patterns rather than form-meaning pairings. The slots in these patterns are characterised by either morpho-syntactic or semantic categories.

The final dimension of interest concerns the \textit{operationality} of the resulting grammars. While all models included in this paper are operational in the sense that they have been computationally implemented,  not all of them yield grammars that support language comprehension and production. In particular, the models based on ECG \citep{chang2008constructing} and  CCG \citep{kwiatkowski2010inducing,kwiatkowski2011lexical,kwiatkowski2012probabilistic,artzi2013weakly,  abend2017bootstrapping} only support language comprehension. The models by \citet{dunn2017computational,dunn2018modeling,dunn2019frequency,dunn2022exposure,dunn2023exploring}, \citet{marti2021discover} and \citet{dunn2021learned} support neither language comprehension nor production. 

A related strand of research that is worth mentioning but falls outside the scope of this paper as defined by the inclusion criteria concerns recent cross-overs between construction grammar and large language models (LLMs). While these approaches share the idea that grammatical patterns can carry non-compositional meaning, they do not (aim to) provide a model for learning construction grammars. As such, \citet{madabushi2020cxgbert} introduce a methodology for finetuning LLMs with the goal of distinguishing between instances of different constructions and \citet{weissweiler2022better,weissweiler2023construction} investigate through probing studies inhowfar LLMs capture constructional form and meaning. For an overview of work on the intersection of construction grammar and LLMs, we refer the interested reader to \citet{tayyar2024construction}.

From the synthesis presented in this section, we can conclude that almost all challenges involved in computationally learning large-scale, usage-based construction grammars have been addressed in some form in prior research, but that no comprehensive models exist to date. So, what would the ultimate model look like and how can we get there? We hereby put forward a number of milestones that will hopefully serve as a roadmap that can boost and streamline future research efforts.

\paragraph{Representing meaning} 
Human languages are constructed by children through interactions with their caregivers. These interactions are meaningful, intentional and situationally grounded. Faithful models of usage-based language acquisition should therefore not rely on direct access to the meaning representations of observed utterances, but implement the process of constructing hypotheses about the intended meaning of the utterances based on the situational context in which they are observed. This entails that the meaning representations need to be composable based on an inventory of pre-linguistic or previously acquired cognitive operations, and be evaluable with respect to a situational context. This process is referred to as \textit{intention reading} in the psycholinguistic literature \citep{tomasello2003constructing}. Initial computational operationalisations of this process have been provided by, amongst others, \citet{spranger2012open}, \citet{pauw2013size} and \citet{nevens2022language}. Semantically annotated corpora are definitely necessary as a scaffold towards the development of large-scale models of construction grammar learning, but cannot be the end point. After all, it is impossible to define, and therefore annotate, meaning independently of contextualised communicative intentions. Ideally, yet probably too ambitious for the short term, the intention reading process should be operationalised using empirically motivated primitive cognitive operations in real-world situations. This will require substantial research efforts into better understanding the building blocks of human cognition, and substantial investments in the development of environments that faithfully simulate the conditions under which human language are acquired, e.g. through the use of virtual reality. 

\paragraph{Representing form} 
The ultimate model represents utterances in their natural form, i.e. as unsegmented sound waves along with data streams that capture information from other modalities, including eye gaze, facial expression and gesture. This sharply contrasts with how form is handled in current models, where all forms are represented as segmented strings of words, graphemes or phonemes. On the one hand, this entails that the learning of grammars is limited by segmentation choices that have been made upfront and independently of the learning process. On the other hand, these representations leave aside a wealth of potentially meaningful information, for example conveyed through prosody or posture. One important innovation in this direction would be to design algorithms that are able to compare and generalise over unsegmented speech signals on the utterance level. The design of such algorithms could start from prior work on the cross-situational learning of auditory vocabularies from semantically grounded speech data \citep{tenbosch2009computational,ons2014fast,renkens2017automatic,wang2022bottleneck} and extend the techniques that are used, so that they can identify patterns above the word level.

\paragraph{Representing constructions}
The ultimate model captures all linguistic knowledge that a language user needs to comprehend and formulate utterances in the form of acquired form-meaning mappings (constructions). This means that no grammar rules, system of categories, or other linguistic structures are predefined. The model can only rely on general strategies to construct form-meaning mappings, combine them, and generalise over them. The grammar should be able to capture form-meaning mappings of varying degrees of abstraction, so that meaning representations that do not mirror the compositionality of the morpho-syntactic structures that express them can also be modelled. The processing engine that performs construction-based language comprehension and production should be able to combine the information captured in large numbers of constructions, so that morpho-syntactically and/or semantically complex utterances can be handled. Indeed, such utterances typically instantiate a wide variety of constructions. Finally, constructions should be able to incorporate sequential word order patterns, agreement patterns or a combination of both. It is therefore important to use a framework that strongly adheres to these basic principles of construction grammar, such as Fluid Construction Grammar \citep{steels2004constructivist,vantrijp2022fcg,beuls2023fluid,beuls2025construction}. 

\paragraph{Learning constructions}
If the input to the language learning process consists of situationally grounded, unsegmented, multi-modal observations of utterances, the first constructions that are learnt can only be holistic pairings between observed utterances and hypotheses about their intended meaning \citep{tomasello2003constructing}. More general constructions can later be distilled as generalisations over both the form and meaning sides of previously acquired constructions with respect to novel observations. This generalisation process requires access to general syntactico-semantic generalisation algorithms. Initial prototypes of such algorithms have been presented by \citet{vaneecke2018generalisation}, \citet{nevens2022language} and \citet{doumen2023modelling}. Apart from improving these algorithms to learn more modular grammars that can, for example, elegantly handle recursive patterns, a crucial target in this direction concerns the design of algorithms that can learn agreement relations on an abstract level, i.e. by expressing congruence without referring to specific categories. Initial efforts in this direction have been presented by \citet{beuls2011simulating}, \citet{vantrijp2012multilevel} and \citet{beuls2013agent}. 

\paragraph{Language-independent learning}
Most prior experiments focus on learning construction grammars based on English data. As a consequence, the resulting models primarily focus on constructions that map between word-order patterns and the meaning they convey. Aspects of meaning expressed through morphology or agreement marking are often not included in the models. While this approach might work for English to a certain extent, it does not generalise to morphology-rich languages, where much of the function of word order might be taken up by marking strategies. It is important to develop learning mechanisms that are applicable to any language, both from a theoretical and from a practical perspective. Theoretically, it would confirm the constructionist idea that all languages can be modelled through constructions. Practically, it would facilitate the development of construction-based language technology applications for a wide variety of languages. 

\paragraph{Scaling up}
A final, more general, criterion concerns the scale of the experiments. As the goal is to learn linguistic capacities from scratch in a human-like manner, it is unavoidable that the language learning process will need to start in very concrete, fully grounded, domain-specific environments. As increasingly more constructions with increasingly higher degrees of abstraction are learnt, the tasks and environments can gradually become more complex, abstract and domain-general. It is important to keep in mind that the ultimate model learns in an incremental fashion, generalises over domains and tasks, and remains forever adaptive to changes in the tasks and environments. While it is natural, and even necessary, to focus on specific parts of the challenge, it is crucial to keep in mind that one day these individual experiments will need to come together.

\section{Conclusion}
\label{Conclusion}

The aim of this paper was to provide an overview of prior work concerning computational models of construction grammar learning, to identify gaps in the state of the art and to propose a perspective on the future of the field. We have first described and compared a wide variety of existing models, and have then synthesised the state of the art with a special focus on the aim of operationalising usage-based construction on a large scale. Finally, we have formulated a number of milestones that can serve as a roadmap towards the development of scalable, language-independent and adaptive techniques for learning construction grammars in a usage-based fashion. 

% ideaal model
We have argued that a comprehensive model of construction grammar learning should learn from meaningful, intentional and situationally grounded communicative interactions. As such, meaning representations need to be actively constructed based on the situational context. Form representations should be free from preprocessing artefacts and should not neglect multi-modal information. Bidirectional construction grammars should emerge as a result of applying general learning strategies and they should cover the full range of linguistic phenomena that occur in the world's languages. We sincerely hope that our synthesis of prior literature  and prospective roadmap can help to boost progress in this area of research, streamline efforts undertaken in different research traditions, and bring us closer to language technologies that can learn to use language in a truly natural, human-like manner.

% relevantie en applications

\bibliographystyle{apalike}

\end{document}